\documentclass[10pt,twocolumn,letterpaper]{article}

\usepackage{iccv}
\usepackage{times}
\usepackage{epsfig}
\usepackage{graphicx}
\usepackage{amsmath}
\usepackage{amssymb}
\usepackage{verbatim}

\usepackage{multirow}
\usepackage{multicol}
\newcommand{\tabincell}[2]{\begin{tabular}{@{}#1@{}}#2\end{tabular}}
\usepackage{color}

\usepackage[pagebackref=true,breaklinks=true,letterpaper=true,colorlinks,bookmarks=false]{hyperref}

\iccvfinalcopy 

\def\iccvPaperID{3295} 

\ificcvfinal\fi
\begin{document}

\title{Multi-Frame Content Integration with a Spatio-Temporal Attention \\ Mechanism for Person Video Motion Transfer }

\author{Kun Cheng\\
Tsinghua University\\
{\tt\small cheng-k16@mails.tsinghua.edu.cn}
\and
Hao-Zhi Huang\\
AI Lab, Tencent\\
{\tt\small matthzhuang@tencent.com}
\and
Chun Yuan\\
Tsinghua University\\
{\tt\small yuanc@sz.tsinghua.edu.cn}
\and
Lingyiqing Zhou\\
Texas A\&M University\\
{\tt\small zlyq11@email.tamu.edu}
\and
Wei Liu\\
Tencent\\
{\tt\small wl2223@columbia.edu}
}

\maketitle

\begin{abstract}
\vspace{-5pt}
Existing person video generation methods either lack the flexibility in controlling both the appearance and motion, or fail to preserve detailed appearance and temporal consistency.
In this paper, we tackle the problem of motion transfer for generating person videos, which provides controls on both the appearance and the motion. Specifically, we transfer the motion of one person in a target video to another person in a source video, while preserving the appearance of the source person. 
Besides only relying on one source frame as the existing state-of-the-art methods, our proposed method integrates information from multiple source frames based on a spatio-temporal attention mechanism to preserve rich appearance details. 
In addition to a spatial discriminator employed for encouraging the frame-level fidelity, a multi-range temporal discriminator is adopted to enforce the generated video to resemble temporal dynamics of a real video in various time ranges. 
A challenging real-world dataset, which contains about 500 dancing video clips with complex and unpredictable motions, is collected for the training and testing.
Extensive experiments show that the proposed method can produce more photo-realistic and temporally consistent person videos than previous methods. As our method decomposes the syntheses of the foreground and background into two branches, a flexible background substitution application can also be achieved.

\end{abstract}

\let\thefootnote\relax\footnotetext{This work was done during Kun's internship at Tencent.}


\begin{figure*}
    \centering
    \setlength{\tabcolsep}{0.5mm}
    \includegraphics[width=0.95\textwidth]{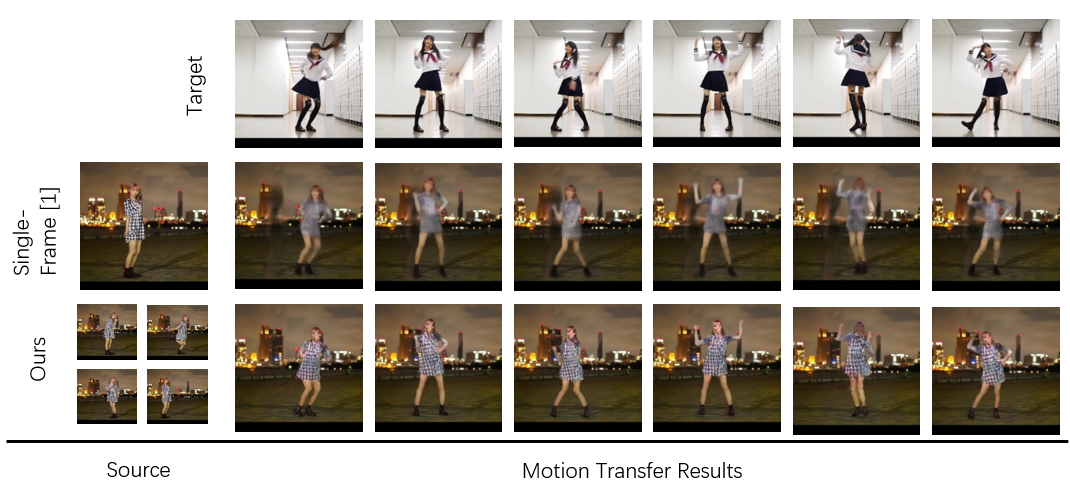}
    \vspace{-5pt}
    \caption{Given a source video and a target video, we transfer the motion from the target to the source. Our method utilize a multi-frame content integration mechanism to combine the information from multiple source frames. In addition, spatio-temporal adversarial losses are adopted to enforce our generated videos to resemble the real ones. Compared to the state-of-the-art single-image model~\cite{balakrishnan2018synthesizing}, our method produces more photo-realistic and temporally consistent results.
    }
    \label{fig:demo}
    \vspace{-12pt}
\end{figure*}

\vspace{-10pt}
\section{Introduction} \label{sec:intro}
\vspace{-3pt}
Person video generation is a cutting-edge topic with plenty of essential applications. It can be used to generate synthetic training data for high-level vision tasks~ \cite{newell2017associative, guler2018densepose, kocabas2018multiposenet,liu2018pose,wei2017person,feichtenhofer2016convolutional, simonyan2014two}, or helping the development of a more powerful video editing tool. 
Current methods for person video generation can be roughly classified into three categories: unconditional video generation, video prediction, and motion transfer.
Unconditional person video generation, which focuses on mapping one or multiple 1D latent vectors to a person video~\cite{tulyakov2017mocogan}, relies on the 1D latent vectors to provide both the appearance and the motion information. 
After training, unconditional video generation models can produce various videos by sampling random latent vector sequences. However, unconditional video generation does not have the flexibility to control the exact motion of the generated video. 
In terms of person video prediction, people are striving to develop methods predicting proceeding frames conditioned on the preceding  frames~\cite{villegas2017decomposing, villegas2017learning}. Person video prediction generally divides the problem into two stages. The first stage concentrates on predicting a pose sequence given the past one, while the second stage turns to generating a person video based on a given pose sequence. 
The second stage is indeed tackling the motion transfer problem. However, the proposed methods for video prediction pay most of the attention to the first stage, while lacking a thorough consideration on preserving appearance details and temporal consistency during the second stage.

As shown in Fig.~\ref{fig:demo}, in this paper, we devote our efforts to the problem of person video motion transfer, which aims at transferring the motion of one person in the target video to another person in the source video, while preserving the appearance of the source. 
Unlike unconditional video generation or video prediction, we obtain a full control of the exact motion in the generated video by assigning a target video with an ideal motion sequence. 
Although there are plenty of approaches striving to solve the pose transfer problem for a single image~\cite{ma2017pose, ma2018disentangled, balakrishnan2018synthesizing, neverova2018dense, pumarola2018unsupervised, siarohin2018deformable}, directly applying them to videos will introduce severe flicker artifacts when the motions are complex and unpredictable. Also, the appearance information from a single image is insufficient for synthesizing an image with a vastly different target pose. Some recent methods~\cite{wang2018video, chan2018everybody} try to narrow this gap by training a video-to-video translator with a fixed-range temporal discriminator to ensure realistic temporal dynamics. In fact, the existing video-to-video methods are to learn a manifold of all the source frames, and then map the target pose sequence to a trajectory on the manifold to generate a new video with the motion from the target and the appearance from the source. Thus, they require to train an individual model for each source video. In addition, both the background and the foreground of the generated video are bound together to be consistent with the source video, without the flexibility of background substitutions.

\vspace{-2pt}
On the contrary, we strive to provide a solution for the task of general person video motion transfer, which can transfer complex and unpredictable motions between any pair of person videos by a single model. 
Instead of modeling a manifold of all the source frames for a single person, our model is capable of inferring the appearance of an arbitrary person in the target pose given only a few source frames.
Besides only counting on a single frame to provide the appearance information as the previous methods~\cite{ma2017pose, ma2018disentangled, balakrishnan2018synthesizing, neverova2018dense, pumarola2018unsupervised, siarohin2018deformable}, we propose a multi-frame content integration strategy to extract rich appearance information from the source video. 
The multi-frame content integration process is guided by a spatio-temporal attention mechanism. With the divided foreground and background branches, our method can also accomplish the background substitution task in a breeze.
In addition to a traditional spatial discriminator, which is adopted to enrich the frame-level photo-realistic details, a multi-range temporal discriminator is proposed to encourage the generated video to resemble the temporal dynamics of a real video over different time ranges. 
To evaluate the model performance for general person video motion transfer, we collect a dataset with around 500 dancing video clips, namely \emph{Dance-500}, which contains complex and unpredictable motions in the wild. The proposed dataset will be made publicly available for the research purpose.
Through quantitative and qualitative comparisons to previous methods, we demonstrate that our proposed method can generate more realistic motion transfer results in terms of both the spatial and temporal domains. 

In summary, the main contributions of this paper are four-fold. i) We propose a flexible motion transfer pipeline, which can be adopted to decompose and recombine the foreground, background and motion information. ii) We introduce a multi-frame content integration module based on a spatio-temporal attention mechanism to produce sharper foreground details and more natural background completion results. iii) We propose a multi-range temporal discriminator to encourage the motion transfer results to be more temporally consistent. 
iv) We collect a challenging person video dataset with complex and unpredictable motions in wild scenes, which is more general for training and evaluating a person video motion transfer model.

\vspace{-5pt}
\section{Related Work} \label{sec:related}
\vspace{-3pt}

\begin{figure*}
    \centering
    \includegraphics[width=0.9\linewidth]{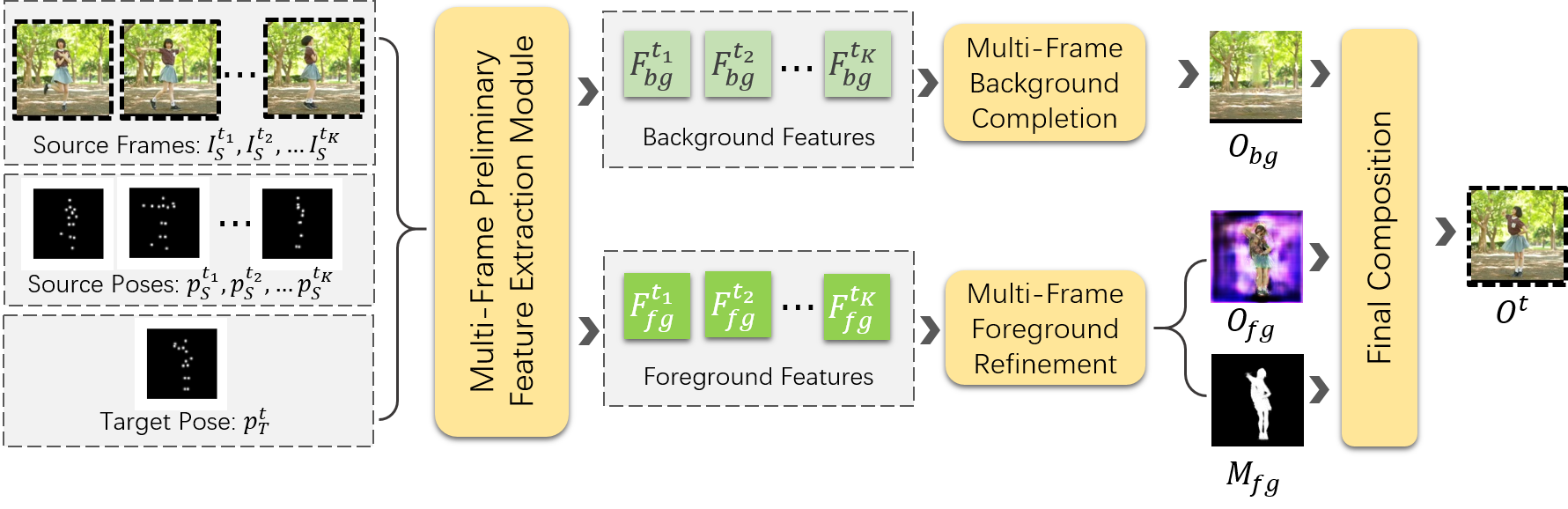}
    \vspace{-5pt}
    \caption{The pipeline of our method. For each time step $t$, our model feeds the $K$ source frames, $K$ source poses and a target pose to a multi-frame preliminary feature extraction module. The computed preliminary features of the foreground and the background are processed separately to produce a synthetic background $O_\text{bg}$, a synthetic foreground $O_\text{fg}$ and a corresponding foreground mask $M_\text{fg}$. Finally, we composite the output frame $O^t$ by combining $O_\text{fg}$ and $O_\text{bg}$ together under the guidance of $M_\text{fg}$.
    }
    \label{fig:Pipeline}
    \vspace{-10pt}
\end{figure*}

\paragraph{Video Prediction.}
Most of the state-of-the-art methods decompose the generation problem into two stages, including a pose sequence prediction stage and a pose guided person video generation stage. The second stage indeed has the same target with motion transfer. In the second stage, Villegas \etal~\cite{villegas2017decomposing, villegas2017learning} and Yang \etal~\cite{yang2018pose} exploit a simple encoder-decoder structure to combine the information of a single source image and a predicted motion code to produce a person image in the target pose. These methods only consider image-level adversarial losses, making the generated results have obvious flicker artifacts.
Recently, Zhao \etal~\cite{zhao2018learning} propose to use an additional refinement stage to post-process the generated video to attain better temporal consistency. At the second stage of Zhao \etal's method, a conditional global temporal discriminator is employed to provide an adversarial loss to train the global refinement network with computationally heavy 3D convolutions. 
However, Zhao \etal's temporal refinement is an individual post-processing stage which is not end-to-end trainable and not practical for handling live streaming videos in a frame-by-frame way.

\vspace{-15pt}
\paragraph{Person Image Generation.}
A wide range of methods have been proposed to generate a realistic person image based on a given pose. 
Ma \etal~\cite{ma2017pose} transfer the human pose of one image to another in a coarse to fine manner. Afterwards, Ma \etal~\cite{ma2018disentangled} propose to disentangle foreground, background and pose features for improving the synthesis quality, which also makes the generation process more controllable. Meanwhile, Balakrishnan \etal~\cite{balakrishnan2018synthesizing} explicitly utilize pose estimation result to split and transform different body parts accordingly, which generates impressive results. However, directly applying the above methods to person video motion transfer frame by frame leads to obvious flicker artifacts, due to the lacking of temporal consistency.

\vspace{-15pt}
\paragraph{Video-to-Video Translation.}
Wang \etal~\cite{wang2018video} propose a general video-to-video translation method, which can be applied to translate a pose sequence into a person dancing video. Chan \etal~\cite{chan2018everybody} propose a more complete system for translating a pose sequence into a person dancing video. Both of the above methods contain a local temporal discriminator, taking 2 or 3 frames as input, to enforce temporal consistency.
Although impressive results are generated by these two methods, they need to train a new model for each source video, lacking the flexibility of applying the trained model to an arbitrary individual. Person video motion transfer problem can also be regarded as a more challenging video-to-video translation problem with two input videos offering messages in different aspects.


\begin{figure*}[t]
    \centering
    \includegraphics[width=0.85\linewidth]{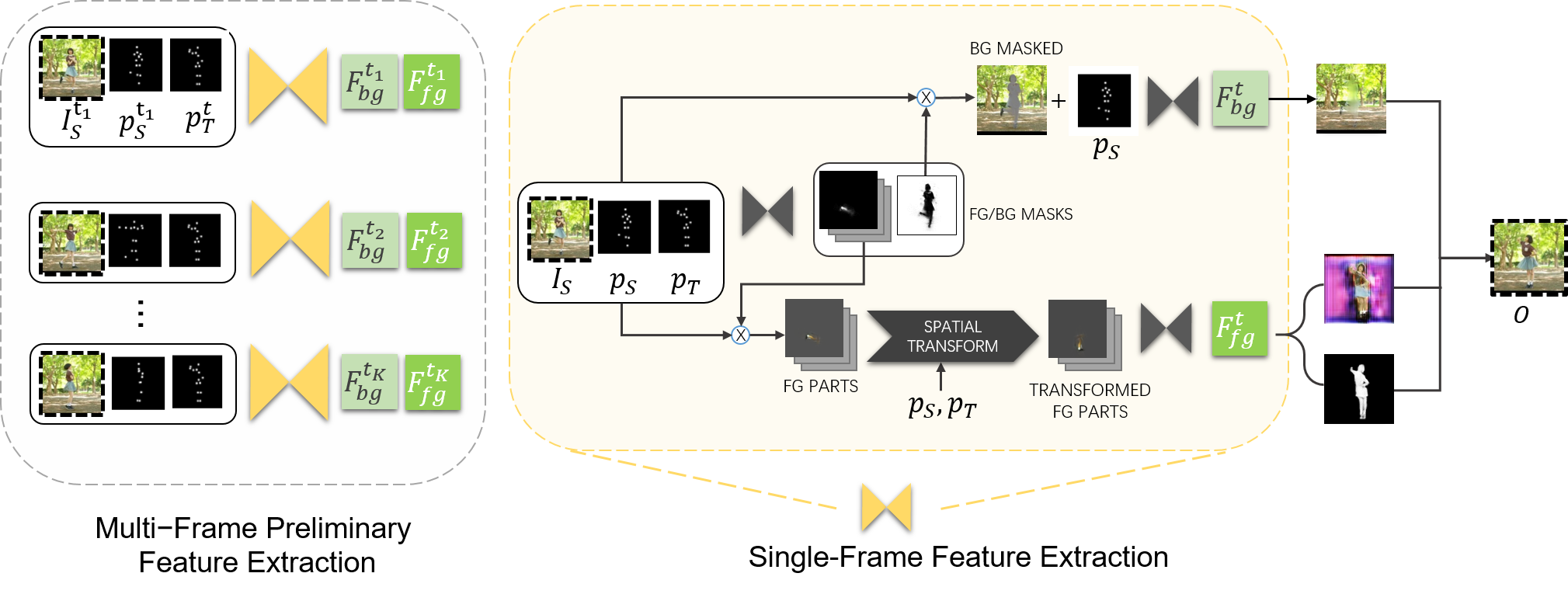}
    \vspace{-5pt}
    \caption{Multi-frame preliminary feature extraction. We take foreground and background features at the second-to-last layer of the single-frame baseline~\cite{balakrishnan2018synthesizing} as the preliminary features. The single-frame baseline contains two branches for the foreground and the background respectively.
    While the background branch learns to accomplish image inpainting to produce a complete background, the foreground branch firstly transfers different body parts using simple affine transforms and then integrates all the parts together to produce a refined foreground. Finally, the foreground and background are combined together according to a predicted foreground mask. 
    }
    \label{fig:Extraction}
    \vspace{-10pt}
\end{figure*}

\vspace{-5pt}
\section{Method} \label{sec:method}
\vspace{-3pt}
We address the problem of person video motion transfer. Here a person video means a video contains $N$ frames {\small{$V=\{I^1, I^2, ..., I^N\}$}} , in which a person conducts whole-body actions like dancing. To simplify the problem, we assume that both the camera and the background are static, which is already a very challenging setting remaining unsolved. Given a source person video $V_S$ and a target person video $V_T$, person video motion transfer aims at transferring the motion of $V_T$ to $V_S$, while preserving the appearance of $V_S$.  This gives an explicit control of both the motion and the appearance of the generated video $V_O$.
A pretrained 2D pose estimator~\cite{fang2017rmpe, xiu2018poseflow} is used to extract motion information in the form of a pose sequence {\small{$P=\{p^1, p^2, ..., p^N\}$}} from a video. Each pose $p^t$ in the $t$-th frame is represented by a $M$-channel heatmap, where {\small{$M=14$}} denotes the number of different keypoints~\cite{balakrishnan2018synthesizing}. 
We denote the pose sequences for the source and target videos as $P_S$ and $P_T$, respectively. 
More advanced pose estimators can be used in the future to acquire more accurate pose information. 

Similar as the single-image pose transfer method proposed by Balakrishnan~\etal~\cite{balakrishnan2018synthesizing}, we handle the synthesis of the foreground and the background separately.  However, instead of conducting pose transfer given only one source image and one target image, we introduce a multi-frame content integration mechanism with a spatio-temporal adversarial training strategy to improve the realism of the results in both the spatial and the temporal aspects.
Specifically, as shown in Fig.~\ref{fig:Pipeline}, our person video motion transfer network $G$ contains four main components, including a multi-frame preliminary feature extraction module, a multi-frame foreground refinement module, a multi-frame background completion module, and a final composition module. We solve the person video motion transfer problem in a frame-by-frame way.  For each time step $t$, our transfer network $G$ takes $K$ source frames, the corresponding $K$ source poses and the target pose as input, generating an output frame {\small$O^t = G(I_\text{S}^{t_1},I_\text{S}^{t_2},...,I_\text{S}^{t_K},p_\text{S}^{t_1},p_\text{S}^{t_2},...,p_\text{S}^{t_K},p_T^t)$}.

The $K$ source frames are randomly chosen from the source video, and fixed thereafter when dealing with the same source video. In this paper, we mainly focus on demonstrating that exploiting multiple source frames sampled in a simple random manner has already brought a significant improvement. A more sophisticated strategy for choosing the keyframes will be left for the future work. 
Based on a single-frame baseline~\cite{balakrishnan2018synthesizing} for extracting preliminary features, we integrate multiple source frames under the guidance of a spatio-temporal attention map to provide more content information, which alleviates the problem of occlusions for both the foreground and background. During the training process, we enforce the network $G$ to generate temporally consistent results with more realistic details by coordinating the training of a series of consecutive frames with spatio-temporal discriminators. 

\vspace{-3pt}
\subsection{Multi-Frame Preliminary Feature Extraction}
\vspace{-2pt}
To extract rich information from the source video, we propose a multi-frame preliminary feature extraction module. 
The preliminary features will be combined together by a multi-frame content integration strategy in the subsequent stages.
We employ a pretrained single-frame pose transfer model \cite{balakrishnan2018synthesizing} to extract preliminary features for each source frame individually, as shown in Fig.~\ref{fig:Extraction}. We adopt \cite{balakrishnan2018synthesizing} as the feature extractor for the impressive visual quality of the generated results and the capability of handling the foreground and background features separately. For each source frame {\small{$I_S^{t_k}$}}, we extract the foreground features {\small{$F_\text{fg}^{t_k}$}} and the background features {\small{$F_\text{bg}^{t_k}$}} at the second-to-last layer of the single-frame model as the preliminary features. Here, {\small{$\{I_S^{t_k} | 1 \leq t_k \leq N, k=1..K\}$}} are {\small{$K$}} frames randomly chosen from the source video. In our experiment, we set {\small{$K=4$}} to balance the computational cost and the generation quality. Both {\small{$F_\text{fg}^{t_k}$}} and {\small{$F_\text{bg}^{t_k}$}} are of size {\small{${C\times H \times W}$}}, where {\small{$C=64$}} and {\small{$H=W=256$}}. We choose the features in the second-to-last layer as the preliminary features instead of the generated images of the single-frame model to reserve more information for the subsequent synthesis modules.

\begin{figure*}[t]
    \centering
    \includegraphics[width=0.9\linewidth]{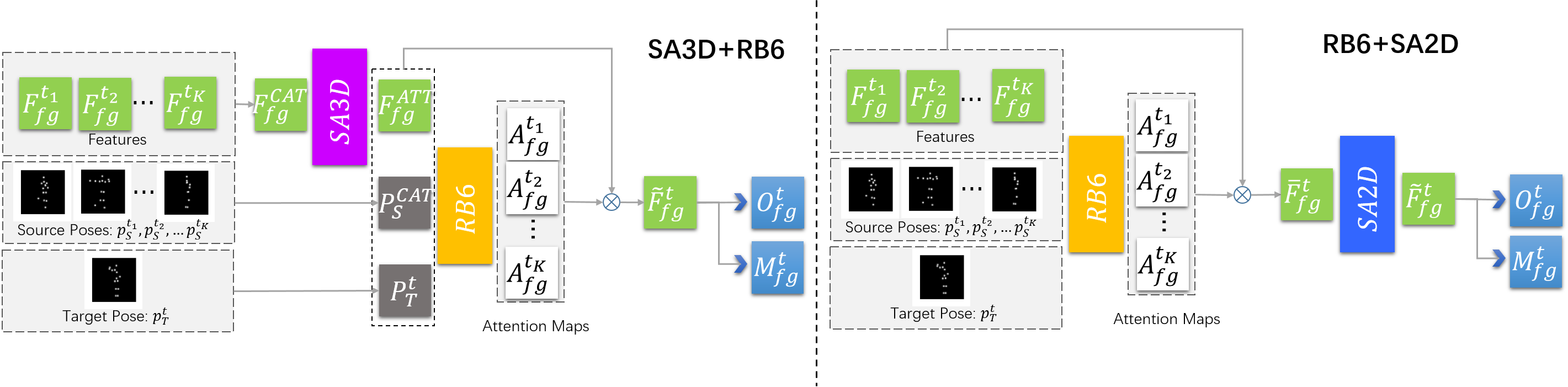}
    \caption{Multi-frame foreground refinement module. The inputs of this module includes the preliminary foreground features, the source poses and the target pose. The structures of two variants ``SA3D+RB6'' and ``RB6+SA2D'' are shown in this figure. 
    }
    \label{fig:foreground}
    \vspace{-10pt}
\end{figure*}

\vspace{-3pt}
\subsection{Multi-Frame Foreground Refinement}\label{sec:fg_refine}
\vspace{-2pt}
In the case of single-frame pose transfer, the quality of the synthetic foreground heavily relies on the source frame choice. For example, a source frame containing the back-view of a person will generate blurry results when given a front-view target pose. The incompleteness of the information from a single view also leads to unstable synthetic results, which will intensify the temporal inconsistency in the generated video.
In this paper, we propose a multi-frame foreground refinement module, which integrates the preliminary foreground features from $K$ source frames to generate a synthetic foreground with higher visual quality. For each time step $t$, the preliminary foreground features $\{F_\text{fg}^{t_1},F_\text{fg}^{t_2},...,F_\text{fg}^{t_K}\}$ from $K$ source frames is fed to a feature fusion module to produce a fused feature map $\widetilde{F}_\text{fg}^t$. On the basis of $\widetilde{F}_\text{fg}^t$, a prediction module is used to generate a synthetic foreground image $O_\text{fg}^t$ and a foreground mask $M_\text{fg}^t$.  The prediction module is a {\small{$3\times3$}} convolutional layer with tanh as the activation layer.

We investigate several feature fusion modules. The most simple and intuitive ones are adding average or max pooling layers to fuse the preliminary features from different source frames. To further exploit the multi-frame information, we explore three variants based on the spatio-temporal attention mechanism.
The inputs of all three variants are the same, including the preliminary foreground features, the source poses and the target pose.
The first variant is named ``RB6'', which takes the concatenation of all the inputs and exploits six residual blocks to compute a ${K\times H\times W}$ spatio-attention map. The spatio-temporal attention map can be regarded as $K$ spatial attention maps of size ${1\times H\times W}$ for the $K$ source frames. 
Then, the fused foreground feature map $\widetilde{F}_\text{fg}^t$ is calculated as: $\widetilde{F}_\text{fg}^t = \sum_{k=1}^K F_\text{fg}^{t_k} \odot A_\text{fg}^{t_k}$.
Here, $F_\text{fg}^{t_k}$ denotes the preliminary foreground feature map from the source frame $I_S^{t_k}$. $A_\text{fg}^{t_k}$ denotes the corresponding attention map. $\odot$ means the element-wise production. Finally, the fused feature map $\widetilde{F}_\text{fg}^t$ is fed to a prediction module to generate a synthetic foreground image $O_\text{fg}^t$ and a corresponding foreground mask $M_\text{fg}^t$.

The drawback of ``RB6'' is that although the attention map is computed based on spatio-temporal information, the features are fused locally across the temporal domain. To alleviate this problem, we propose two more sophisticated variants ``SA3D+RB6'' and ``RB6+SA2D'' as shown in Fig.~\ref{fig:foreground}. 
``SA3D+RB6'' first exploits a 3D self-attention sub-module ``SA3D'' to fuse all features non-locally across both the spatial and temporal domain, and then a ``RB6'' sub-module is exploited to refine the fused feature. A non-local attention map of size ${( K\times H\times W) \times (K\times H\times W)}$ is computed for the spatial-temporal feature fusion in ``SA3D'', which is different from the spatial self-attention mechanism ``SA2D'' proposed in~\cite{zhang2018self}. ``RB6+SA2D''  first exploits ``RB6'' to fuse the features locally across the temporal domain, and then  ``SA2D'' is exploited to fuse the features non-locally across the spatial domain.  Through experiments, we find that by decomposing the spatial-temporal non-local fusion into two stages, ``RB6+SA2D'' alleviates the computational burden and achieves a similar performance as ``SA3D+RB6''.

\vspace{-3pt}
\subsection{Multi-Frame Background Completion.}
\vspace{-2pt}
Similar to the foreground synthesis, the background completion mission can also benefit from multiple source frames. Since the pose of the person varies from frame to frame, the occluded background area in one frame may be observed in another frame. We can integrate multi-frame information to easily generate a more accurate background completion result.
Similarly, the multi-frame background completion module takes the concatenated preliminary background features $\{F_\text{bg}^{t_1},F_\text{bg}^{t_2},...,F_\text{bg}^{t_K}\}$, source poses $\{p_\text{S}^{t_1},p_\text{S}^{t_2},...,p_\text{S}^{t_K}\}$ and the target pose $p_T^t$ as input.
Then, a fused background feature map $\widetilde{F}_\text{bg}^t$ is calculated similarly as in Section~\ref{sec:fg_refine}.
Finally, $\widetilde{F}_\text{bg}^t$ is fed to a prediction module to generate a synthetic background image $O_\text{bg}^t$.

\vspace{-3pt}
\subsection{Final Composition.}
\vspace{-2pt}
For each time step $t$, given the synthetic foreground image $O_\text{fg}^t$, the synthetic background image $O_\text{bg}^t$ and the foreground mask $M_\text{fg}^t$, the final composite frame should be: {\small$O^t = O_\text{fg}^t \odot M_\text{fg}^t + O_\text{bg}^t \odot (1 - M_\text{fg}^t)$}.

\vspace{-3pt}
\subsection{Spatio-Temporal Adversarial Training}
\vspace{-2pt}
To train the transfer network {\small$G$} to generate photo-realistic and temporally consistent results, we adopt a multi-frame coordinated training strategy with the help of spatio-temporal adversarial losses. During training, given {\small$K$} randomly chosen source frames and {\small$L$} consecutive target frames, the transfer network {\small$G$} generates {\small$L$} corresponding output frames {\small$V_O^{[t-L+1,t]}=\{O^{t-L+1},O^{t-L+2},...,O^{t}\}$} in a frame-by-frame way. On one hand, we encourage that each generated frame to be photo-realistic and contain the source person performing the target pose. On the other hand, we encourage the generated {\small$L$} consecutive frames to resemble the temporal dynamics of the {\small$L$} consecutive target frames.
These two goals are achieved by introducing a combination of a content loss, a spatial adversarial loss and a multi-range temporal adversarial loss.

\vspace{-15pt}
\paragraph{Content Loss.}
To enable supervised training, we use the same video to provide the source frames and the target frames for the training phase. We make sure that the source frames and target frames have no overlap. After training, for an arbitrary source video, we can choose an arbitrary target video to provide the target pose sequence. Under the supervised training setting, we know that a generated frame $O^t$ should be equal to the target frame $I_T^t$. Thus, the most straight forward loss is the mean square errors (MSEs) between the generated frames and the ground-truth frames: $\mathcal{L}_\text{MSE} = \sum_t \big\| O^t - I_T^t \big\|_2^2$.
However, this loss function tends to produce blurry results, which is because the generator is learning to agree with as many solutions as possible and finally comes to an average solution. 
To add more details, a perceptual loss suggested in~\cite{johnson2016perceptual, ledig2017photo} is also adopted:$\mathcal{L}_\text{VGG} = \sum_t\big\|\phi(O^t)-\phi(I_T^t)\big\|_{1}$.
Here, $\phi$ denotes the features extracted by a pretrained VGG19 model~\cite{simonyan2014very}. In our experiment, we use the features at the conv1\_1, conv2\_1, conv3\_1 and conv4\_1 layers. 
$\mathcal{L}_\text{VGG}$ encourages the generated frame to be close to the ground-truth frame in the semantic feature domain defined by a pre-trained VGG network, which enhances the perceptual similarity.

\vspace{-15pt}
\paragraph{Spatial Adversarial Loss.} 
To encourage each generated frame contains more photo-realistic details, we introduce a spatial adversarial loss. A single-frame conditional discriminator $D_I$ is trained to distinguish the generated frame from the ground-truth frame.
We use LSGAN~\cite{mao2017least} and PatchGAN~\cite{isola2017image} for stable training:
\begin{small} 
\begin{align}
{\small
    \mathcal{L}_\text{GAN, I}=\sum_t {\big\|D_I(I_T^t,p_{T}^{t})\big\|_2^2 + \big\|1 - D_I(O^ t,p_{T}^{ t} )\big\|_2^2}.}
\end{align}
\end{small} 

\vspace{-20pt}
\paragraph{Multi-Range Temporal Adversarial Loss.} Besides the spatial adversarial loss, we also introduce a multi-range temporal adversarial loss to encourage the generated video to resemble the temporal dynamics of a real video. Instead of using only one fixed-range temporal discriminator as in ~\cite{wang2018video}, multiple temporal discriminators 
are trained to verify the temporal consistency over different time ranges. The multi-range temporal adversarial loss is defined as:
\begin{small}
\begin{equation} 
\begin{aligned}
    \mathcal{L}_\text{GAN, V} & =  \sum_n \sum_t {\Big\|D_{V}^{n}(V_T^{[t-n+1,t]},W_T^{[t-n+2,t]})\Big\|}_2^2  \\
    & + {\Big\|1 - D_{V}^{n}(V_O^{[t-n+1,t]},W_T^{[t-n+2,t]})\Big\|}_2^2. 
\end{aligned}
\end{equation}
\end{small} 
Here, $W_T$ denotes an optical flow sequence calculated by FlowNet2~\cite{ilg2017flownet}, which contains the optical flows between every pair of adjacent frames. $D_{V}^n$ denotes a temporal discriminator, which takes every $n$ frames and the optical flows between them as input, and learns to distinguish the generated $n$ consecutive frames from the ground-truth ones.

\vspace{-15pt}
\paragraph{Total Loss.}
Finally, the total training loss is defined as:
\begin{small} 
\begin{equation} 
\begin{aligned}
{\small
    \mathcal{L}_\text{total}  =  \mathcal{L}_\text{MSE} + \lambda_\text{VGG} \mathcal{L}_\text{VGG} + \lambda_\text{GI} \mathcal{L}_\text{GAN, I} + \lambda_\text{GV} \mathcal{L}_\text{GAN, V}
}
\end{aligned}
\end{equation}
\end{small} 
We aim to solve: $\arg \min_{G} \max_{D_I,D_V} \mathcal{L_\text{total}}(G,D_I,D_V)$.
Here, $D_V=\{ D_V^n | n=3,5,7\}$ represents the collection of the temporal discriminators over different time ranges. This objective function can be optimized by alternately updating the discriminators and the generator.

\begin{figure*}
    \centering
    
    \setlength{\tabcolsep}{0.5mm}
    \vspace{0pt}

    \includegraphics[width=0.8\textwidth]{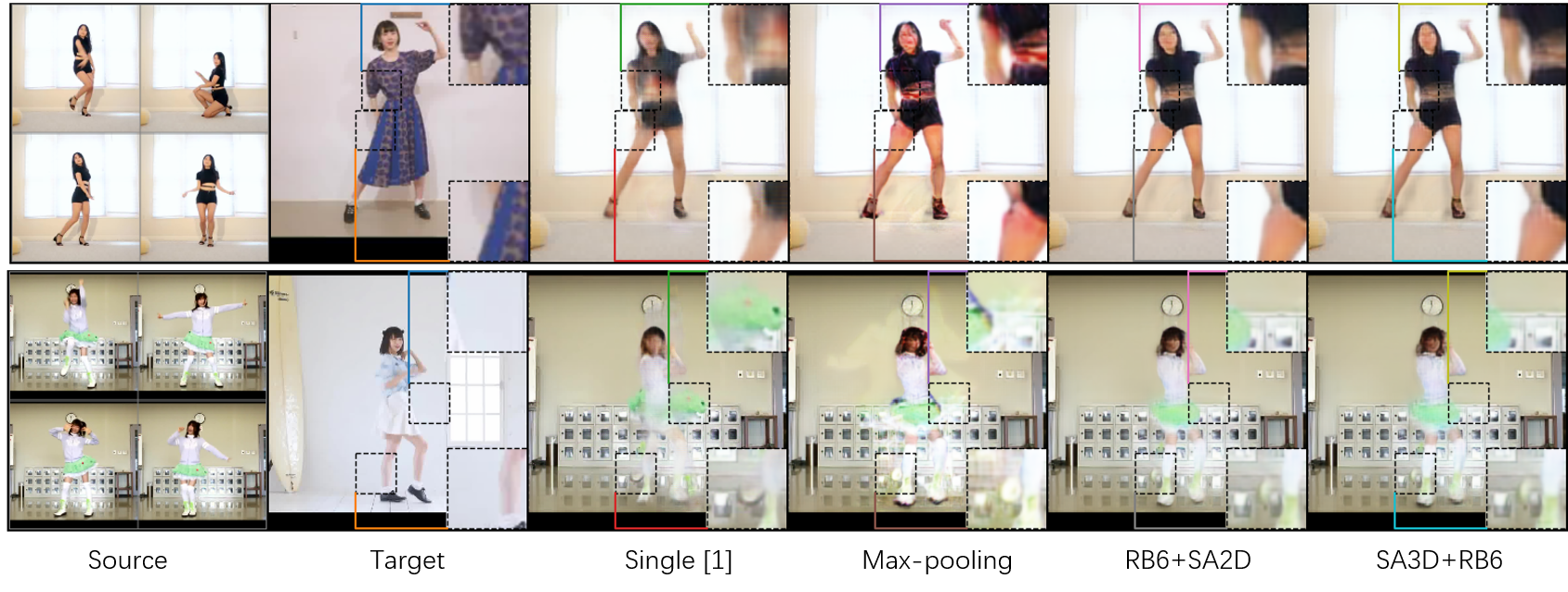}
    \caption{Results on the \emph{Cross-Video} subset. The results of the single-frame method~\cite{balakrishnan2018synthesizing} are blurry. The max pooling variant introduces some sharp details but its colors are strange. ``SA2D'' and ``SA3D'' achieve the best overall performance.
    }
    \label{fig:cross-video}
    \vspace{-5pt}
\end{figure*}

\begin{figure*}
    \centering
    
    \setlength{\tabcolsep}{0.5mm}
    \vspace{0pt}

    \includegraphics[width=0.8\textwidth]{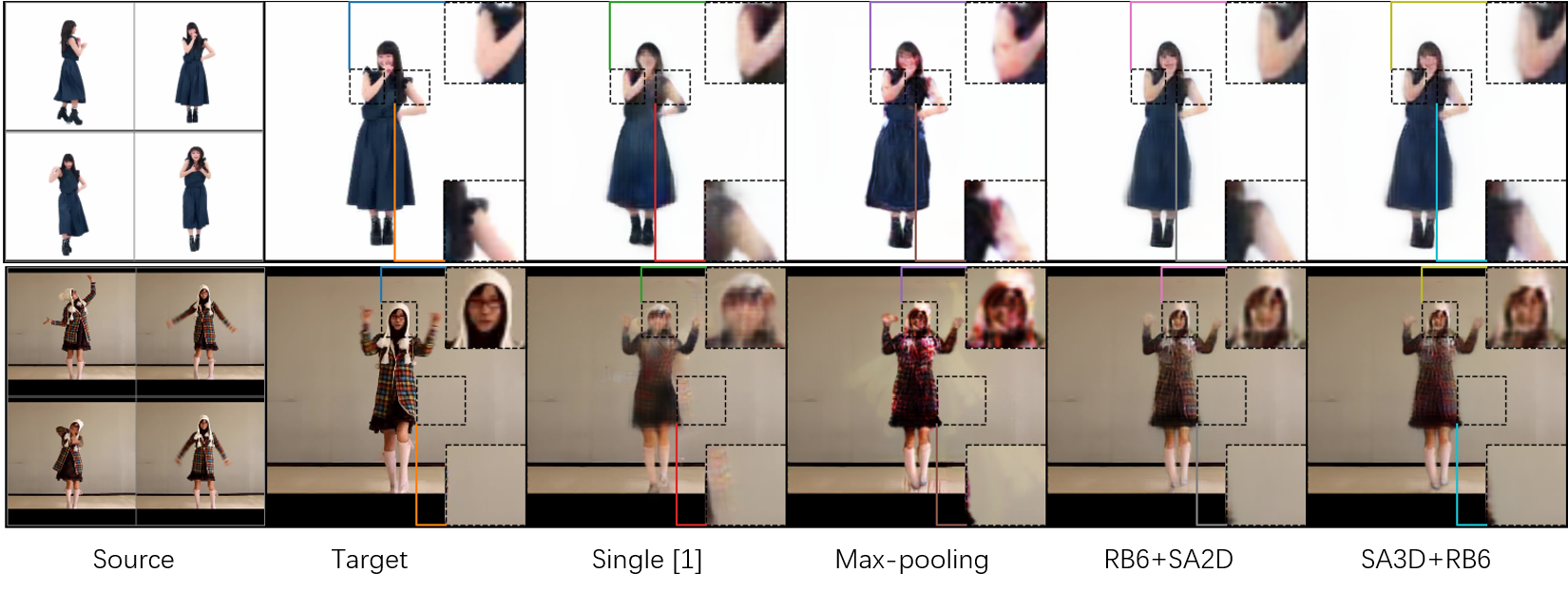}
    \caption{Results on the \emph{Same-Video} subset. The results of the single-frame method~\cite{balakrishnan2018synthesizing} are blurry. The max pooling varaint introduces some sharp details but its colors are strange. ``SA2D'' and ``SA3D'' achieve the best overall performance. 
    }
    \label{fig:inside-video}
    \vspace{-10pt}
\end{figure*}

\begin{figure*}
    \centering
    
    \setlength{\tabcolsep}{0.5mm}
    \vspace{0pt}

    \includegraphics[width=0.95\textwidth]{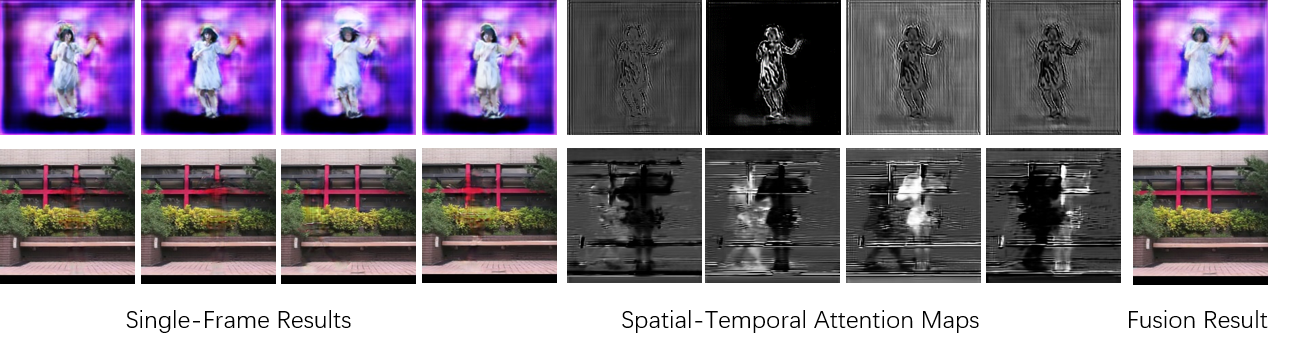}
    \vspace{-5pt}
    \caption{Intermediate results of multi-frame content integration. The 1st row is the foreground result, the 2nd row is the background result. The attention maps indicate the ``comfort zones'' of different source frames. 
    }
    \vspace{-5pt}
    \label{fig:multi_frame}
\end{figure*}

\begin{figure}
    \centering
    
    \vspace{0pt}
    \includegraphics[width=0.45\textwidth]{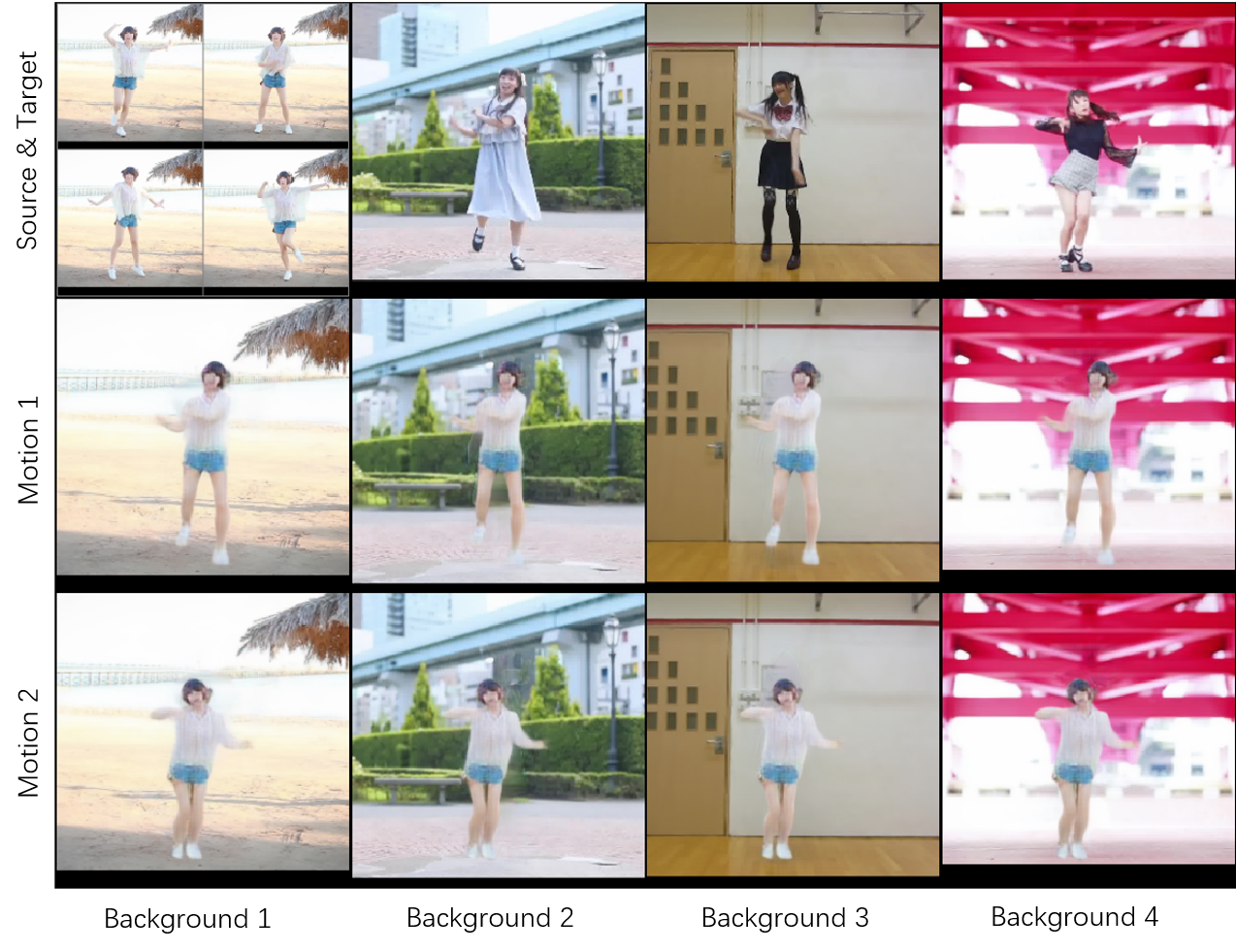}
    \caption{ Examples for background substitution. Motion 1 and 2 comes from the 1st and 3rd target person, respectively. Person A can do actions like person B in front of different backgrounds.
    }
    \label{fig:bg-sub}
    \vspace{-10pt}
\end{figure}

\vspace{-5pt}
\section{Experiment} \label{sec:experiment}
\vspace{-3pt}

\paragraph{Dataset.}
We construct a novel video dataset, \emph{Dance-500}, for training and evaluating models for general person video motion transfer. 
We collect 163 dance videos from Youtube, which contain either different individuals or the same individual in different clothes. Each video presents a different dance. Both the camera and background are static at most of the time.
The videos in the dataset includes both the front and back sides of a person, fast and large motions, and various foreground and background appearances.  
Each video is randomly cut into 3 to 4 clips lasting for 10 seconds. Finally, we obtain about 500 clips with more than 125,000 frames. 
Finally, we use 400 clips for training, 50 clips for validation, and 50 clips for testing.
The datasets used in~\cite{balakrishnan2018synthesizing, tulyakov2017mocogan} only cover some simple action categories, including golf, yoga and tennis, Tai-Chi, jumping-jack, waving-hands and so on. The motions in theses action categories are monotonous and predictable. 
Compared to the above datasets, our proposed \emph{Dance-500} dataset contains a larger number of videos clips with complex and unpredictable motions in wild scenes.  \emph{Dance-500} provides rich information for training and challenging examples for testing. Some cross-dataset testing results are provided in the supplementary material to demonstrate the superiority of the proposed \emph{Dance-500}.

\vspace{-15pt}
\paragraph{Implementation Details.}
In the preliminary feature extraction module, the single-frame model follows the network architecture of \cite{balakrishnan2018synthesizing}. The structure of the self-attention sub-module is similar as~\cite{zhang2018self}.
The spatial and temporal discriminators consist of six residual blocks following~\cite{wang2018video}.  
Please refer to the supplementary material for the detailed architectures of our networks. 
We adopt the Adam~\cite{kingma2014adam} with lr = $0.0001$ and $(\beta_1,\beta_2) = (0.5,0.999)$ for training. For preliminary feature extraction, we pretrain the single-frame pose transfer model using MSE loss and a single-frame GAN loss for 200,000 iterations with a batch size of 4. The single-frame model is fixed thereafter. Then, we train our full model for 35,000 iterations with a batch size of 2, while each sample in the batch consists of 8 consecutive target frames and 4 random source frames. The snapshot with the best validation performance is recorded as the training goes on.
Each frame is cropped in the size of $256 \times 256$ with the person in the center. 
We set {\small$\{  \lambda_{VGG}, \lambda_{GI}, \lambda_{GV} \} = \{0.2,0.2,0.1\}$} in our experiment.

\begin{table}[]
    \centering
    \resizebox{1.0\columnwidth}{!}{
    \begin{tabular}{|c|c|c|c|c|}
         \hline
         \tabincell{c}{Losses} & VFID  & PSNR  \\ 
         \hline
         MSE  & 10.25 & 22.10   \\
         \hline
         MSE + VGG & 9.50 & 22.83    \\
         \hline
         \tabincell{c}{MSE + VGG + $Fusion_{RB6}$ } & 7.99 & 22.79   \\
         \hline
         \tabincell{c}{MSE + VGG + $Fusion_{RB6}$ + $D_V^3$} & 8.09 & 23.07   \\
         \hline
         \tabincell{c}{MSE + VGG + $Fusion_{RB6}$ + $D_V$} & 7.93 & 22.96   \\
         \hline
         \tabincell{c}{MSE + VGG + $Fusion_{Avg}$ + $D_V$} & 8.06 & 22.83   \\
         \hline
         \tabincell{c}{MSE + VGG + $Fusion_{Max}$ + $D_V$} & \textbf{7.47} & 21.37   \\
         \hline
         
         \tabincell{c}{ MSE + VGG + $Fusion_{SA3D+RB6}$ + $D_V$} & 7.74 &  \textbf{23.25}   \\
         \hline
         \tabincell{c}{ MSE + VGG + $Fusion_{RB6+SA2D}$ + $D_V$} & \textbf{7.57} & \textbf{23.20}   \\
         \hline
    \end{tabular}
    }
    \caption{The quantitative results on the \emph{Same-Video} subset.}
    \vspace{-5pt}
    \label{tab:in_video}
\end{table}

\begin{table}[t]
    \centering
    \resizebox{1.0\columnwidth}{!}{
    \begin{tabular}{c|ccc}
        Methods & Single-Frame & Ours & Not Sure  \\
        \hline
        \tabincell{c}{Overall Fidelity} & 17\% & \textbf{77.5\%} & 5.5\% \\
        \tabincell{c}{Temporal Consistency} & 18.5\% & \textbf{65\%} & 16.5\% \\
    \end{tabular}
    }
    \caption{Human Preference Score. }
    \vspace{-10pt}
    \label{tab:userstudy}
\end{table}

\vspace{-3pt}
\subsection{Quantitative Comparisons to Existing Methods}
\vspace{-2pt}



The VFID~\cite{wang2018video} and PSNR scores are adopted for quantitative comparisons. To compute the VFID score, we first extract the video features by a pretrained video classificatoin network I3D~\cite{carreira2017quo}. Then, the mean $\Tilde{\mu}$ and the covariance matrix $\Tilde{\Sigma}$ of the feature vectors are computed over all videos in a dataset. Finally, the VFID score is calculated as $\|\mu-\Tilde{\mu}\|^2+Tr(\Sigma+\Tilde{\Sigma}-2\sqrt{\Sigma\Tilde{\Sigma}})$. The VFID score measures both the visual quality and the temporal consistency. For the \emph{Same-Video} subset, the ground-truth video is equal to the target video. Thus, we can also directly calculate the PSNR for each generated frame. For the \emph{Cross-Video} subset, the appearances of synthesized video and the target video are totally different, which means PSNR is not available. Also, we can not compute the VFID scores, because besides the motion information, the appearance information of a video also affect the features extracted by I3D~\cite{carreira2017quo}. Thus, we only provide quantitative evaluation on the \emph{Same-Video} subset.

Table.~\ref{tab:in_video} shows the VFID and PSNR scores of different methods on the \emph{Same-Video} subset. The top two scores under each metric are highlighted. 
Comparing the results of ``MSE'' and ``MSE+VGG'' for the single-frame baseline, we know that introducing a VGG loss as a part of the content loss in addition to the MSE loss improves both the frame-level quality and the video-level temporal dynamics. Comparing ``MSE+VGG'' with different ``MSE+VGG+Fusion'' variants, we can observe a significant improvement of the VFID score, which indicates that the multi-frame content integration mechanism greatly benefits the video-level perceptual quality.
Comparing ``RB6'' with ``RB6+$D_V$ '', we can see consistent improvements under both metrics can be achieved by introducing a multi-range temporal discriminator.
Comparing ``RB6+$D_V^3$'' with ``RB6+$D_V$'', we know that although the PSNR of $D_V^3$ is a little bit better than $D_V$, it comes at the cost of sacrificing the overall video-level perceptual quality. 
Among different fusion manners, ``Max'' shows the best VFID score but the worst PSNR, indicating that the max pooling variant introduces some nonsense details to make the results look realistic.
The last two rows show that ``SA3D+RB6'' achieves the best PSNR score, while ``RB6+SA2D'' shows an outstanding performance under both metrics.





We also conduct a user study to compare ``RB6+SA2D'' to the single-frame method~\cite{balakrishnan2018synthesizing}. For each method, we generate 5 videos under the \emph{Same-Video} setting and 5 videos under the \emph{Cross-Video} setting. The results from different methods were shown in a random order for a fair comparison. We asked the user to choose the preferred result according to two criteria. The first criterion is about the overall fidelity, as we ask ``which video looks more realistic''. The second criterion is about the temporal consistency, as we ask ``which video contains less flicker artifacts''. In summary, 20 people aged from 20 to 30 have participated. We report the average human preference scores in Table~\ref{tab:userstudy}, which shows that our method surpasses the state-of-the-art single-frame method by a large margin.

\vspace{-3pt}
\subsection{Qualitative Comparisons to Existing Methods}
\vspace{-2pt}

We adopt two kinds of testing subsets to evaluate different methods: i) A \emph{Cross-Video} subset, where the source and the target come from different videos.  ii) An \emph{Same-Video} subset, where the source and the target come from the same video.
We randomly choose 50 pairs of videos from the testing set to construct each subset. 
Note that for the \emph{Same-Video} subset, we ensure that the source sequence and the target sequence have no overlap. 
Fig.~\ref{fig:cross-video} and Fig.~\ref{fig:inside-video} show some qualitative results on the two testing subsets. 
Obvious blurry artifacts can be observed in the results of the single-frame baseline. The max pooling variant tends to generate strange colors in both the foreground and background. ``RB6+SA2D'' and ``SA3D+RB6'' shows the best overall performances. By integrating the content information from multiple source frames based on a spatio-temporal attention mechanism, our background completion results are more accurate, while the foregrounds preserve more details. Please refer to the supplementary video for a clearer observation. 

To give a further investigation of the multi-frame content integration mechanism, we visualize some intermediate results of the ``RB6+SA2D'' variant in Fig.~\ref{fig:multi_frame}.  Here, only the attention maps generated by the ``RB6'' sub-module are shown due to limited space. Obvious artifacts can be observed in the single-frame pose transfer results. Each single-frame result may produce appealing details in one area but blurry textures in another area. Our proposed method learns to compute a spatio-temporal attention map to locate the ``comfort zones'' for each source frame, and guide the feature fusion process for producing synthetic foreground and background with more accurate details.


\vspace{-3pt}
\subsection{Background Substitution}
\vspace{-2pt}
As our method conduct the foreground refinement and the background completion in two branches, with a slight modification, our pipeline can also be used to accomplish both the motion transfer and the background substitution simultaneously. Specifically, we apply the background completion branch to a third video to offer a new background, while the foreground branch is still working on transferring the target pose to the source person. Fig.~\ref{fig:bg-sub} shows some of the background substitution results. Please refer to the supplementary video for more results.

\vspace{-5pt}
\section{Conclusion} \label{sec:conclusion}
\vspace{-3pt}
In this paper, we proposed a person video motion transfer approach, which leverages a multi-frame source content integration mechanism and a spatio-temporal adversarial training procedure to encourage the transfer network to generate temporally consistent videos with more photo-realistic details. The extensive experiments on a proposed challenging dataset \emph{Dance-500} demonstrate that our approach outperforms previous methods in terms of both the single-frame quality and the video temporal consistency.

{\small
\bibliographystyle{ieee}
\bibliography{egbib}

\begin{thebibliography}{10}\itemsep=-1pt

\bibitem{balakrishnan2018synthesizing}
G.~Balakrishnan, A.~Zhao, A.~V. Dalca, F.~Durand, and J.~Guttag.
\newblock Synthesizing images of humans in unseen poses.
\newblock In {\em CVPR}, 2018.

\bibitem{carreira2017quo}
J.~Carreira and A.~Zisserman.
\newblock Quo vadis, action recognition? a new model and the kinetics dataset.
\newblock In {\em CVPR}, 2017.

\bibitem{chan2018everybody}
C.~Chan, S.~Ginosar, T.~Zhou, and A.~A. Efros.
\newblock Everybody dance now.
\newblock {\em arXiv preprint arXiv:1808.07371}, 2018.

\bibitem{fang2017rmpe}
H.-S. Fang, S.~Xie, Y.-W. Tai, and C.~Lu.
\newblock {RMPE}: Regional multi-person pose estimation.
\newblock In {\em ICCV}, 2017.

\bibitem{feichtenhofer2016convolutional}
C.~Feichtenhofer, A.~Pinz, and A.~Zisserman.
\newblock Convolutional two-stream network fusion for video action recognition.
\newblock In {\em CVPR}, 2016.

\bibitem{guler2018densepose}
R.~A. G{\"u}ler, N.~Neverova, and I.~Kokkinos.
\newblock Densepose: Dense human pose estimation in the wild.
\newblock In {\em CVPR}, 2018.

\bibitem{ilg2017flownet}
E.~Ilg, N.~Mayer, T.~Saikia, M.~Keuper, A.~Dosovitskiy, and T.~Brox.
\newblock Flownet 2.0: Evolution of optical flow estimation with deep networks.
\newblock In {\em CVPR}, 2017.

\bibitem{isola2017image}
P.~Isola, J.-Y. Zhu, T.~Zhou, and A.~A. Efros.
\newblock Image-to-image translation with conditional adversarial networks.
\newblock In {\em CVPR}, 2017.

\bibitem{johnson2016perceptual}
J.~Johnson, A.~Alahi, and L.~Fei-Fei.
\newblock Perceptual losses for real-time style transfer and super-resolution.
\newblock In {\em ECCV}, 2016.

\bibitem{kingma2014adam}
D.~P. Kingma and J.~Ba.
\newblock Adam: A method for stochastic optimization.
\newblock In {\em ICLR}, 2015.

\bibitem{kocabas2018multiposenet}
M.~Kocabas, S.~Karagoz, and E.~Akbas.
\newblock Multiposenet: Fast multi-person pose estimation using pose residual
  network.
\newblock In {\em ECCV}, 2018.

\bibitem{ledig2017photo}
C.~Ledig, L.~Theis, F.~Husz{\'a}r, J.~Caballero, A.~Cunningham, A.~Acosta,
  A.~P. Aitken, A.~Tejani, J.~Totz, Z.~Wang, et~al.
\newblock Photo-realistic single image super-resolution using a generative
  adversarial network.
\newblock In {\em CVPR}, 2017.

\bibitem{liu2018pose}
J.~Liu, B.~Ni, Y.~Yan, P.~Zhou, S.~Cheng, and J.~Hu.
\newblock Pose transferrable person re-identification.
\newblock In {\em CVPR}, 2018.

\bibitem{ma2017pose}
L.~Ma, X.~Jia, Q.~Sun, B.~Schiele, T.~Tuytelaars, and L.~Van~Gool.
\newblock Pose guided person image generation.
\newblock In {\em NeurIPS}, 2017.

\bibitem{ma2018disentangled}
L.~Ma, Q.~Sun, S.~Georgoulis, L.~Van~Gool, B.~Schiele, and M.~Fritz.
\newblock Disentangled person image generation.
\newblock In {\em CVPR}, 2018.

\bibitem{mao2017least}
X.~Mao, Q.~Li, H.~Xie, R.~Y. Lau, Z.~Wang, and S.~P. Smolley.
\newblock Least squares generative adversarial networks.
\newblock In {\em ICCV}, 2017.

\bibitem{neverova2018dense}
N.~Neverova, R.~Alp~Guler, and I.~Kokkinos.
\newblock Dense pose transfer.
\newblock In {\em ECCV}, 2018.

\bibitem{newell2017associative}
A.~Newell, Z.~Huang, and J.~Deng.
\newblock Associative embedding: End-to-end learning for joint detection and
  grouping.
\newblock In {\em NeurIPS}, 2017.

\bibitem{pumarola2018unsupervised}
A.~Pumarola, A.~Agudo, A.~Sanfeliu, and F.~Moreno-Noguer.
\newblock Unsupervised person image synthesis in arbitrary poses.
\newblock In {\em CVPR}, 2018.

\bibitem{siarohin2018deformable}
A.~Siarohin, E.~Sangineto, S.~Lathuili{\`e}re, and N.~Sebe.
\newblock Deformable gans for pose-based human image generation.
\newblock In {\em CVPR}, 2018.

\bibitem{simonyan2014two}
K.~Simonyan and A.~Zisserman.
\newblock Two-stream convolutional networks for action recognition in videos.
\newblock In {\em NeurIPS}, 2014.

\bibitem{simonyan2014very}
K.~Simonyan and A.~Zisserman.
\newblock Very deep convolutional networks for large-scale image recognition.
\newblock In {\em ICLR}, 2015.

\bibitem{tulyakov2017mocogan}
S.~Tulyakov, M.-Y. Liu, X.~Yang, and J.~Kautz.
\newblock {MoCoGAN}: Decomposing motion and content for video generation.
\newblock In {\em CVPR}, 2018.

\bibitem{villegas2017decomposing}
R.~Villegas, J.~Yang, S.~Hong, X.~Lin, and H.~Lee.
\newblock Decomposing motion and content for natural video sequence prediction.
\newblock In {\em ICLR}, 2017.

\bibitem{villegas2017learning}
R.~Villegas, J.~Yang, Y.~Zou, S.~Sohn, X.~Lin, and H.~Lee.
\newblock Learning to generate long-term future via hierarchical prediction.
\newblock In {\em ICML}, 2017.

\bibitem{wang2018video}
T.-C. Wang, M.-Y. Liu, J.-Y. Zhu, G.~Liu, A.~Tao, J.~Kautz, and B.~Catanzaro.
\newblock Video-to-video synthesis.
\newblock In {\em NeurIPS}, 2018.

\bibitem{wei2017person}
L.~Wei, S.~Zhang, W.~Gao, and Q.~Tian.
\newblock Person transfer gan to bridge domain gap for person
  re-identification.
\newblock In {\em CVPR}, 2018.

\bibitem{xiu2018poseflow}
Y.~Xiu, J.~Li, H.~Wang, Y.~Fang, and C.~Lu.
\newblock {Pose Flow}: Efficient online pose tracking.
\newblock In {\em BMVC}, 2018.

\bibitem{yang2018pose}
C.~Yang, Z.~Wang, X.~Zhu, C.~Huang, J.~Shi, and D.~Lin.
\newblock Pose guided human video generation.
\newblock In {\em ECCV}, 2018.

\bibitem{zhang2018self}
H.~Zhang, I.~Goodfellow, D.~Metaxas, and A.~Odena.
\newblock Self-attention generative adversarial networks.
\newblock {\em arXiv preprint arXiv:1805.08318}, 2018.

\bibitem{zhao2018learning}
L.~Zhao, X.~Peng, Y.~Tian, M.~Kapadia, and J.~Metaxas, Dimitris.
\newblock Learning to forecast and refine residual motion for image-to-video
  generation.
\newblock In {\em ECCV}, 2018.

\end{thebibliography}
}

\end{document}